\documentclass[conference]{IEEEtran}
\IEEEoverridecommandlockouts

\usepackage{cite}
\usepackage{amsmath,amssymb,amsfonts}
\usepackage{algorithmic}
\usepackage{graphicx}
\usepackage{textcomp}
\usepackage{xcolor}
\usepackage{colortbl}
\usepackage{subcaption}
\usepackage{booktabs}
\usepackage{tabularx}
\usepackage{makecell}
\usepackage{hyperref}
\usepackage{float}
\usepackage{listings}
\usepackage{stfloats}

\def\BibTeX{{\rm B\kern-.05em{\sc i\kern-.025em b}\kern-.08em
    T\kern-.1667em\lower.7ex\hbox{E}\kern-.125emX}}

\begin{document}


\title{LLaVul: A Multimodal LLM for Interpretable Vulnerability Reasoning about Source Code}



\author{
    \textbf{Ala Jararweh}$^{*,1,2}$, \textbf{Michael Adams}$^{*,1}$, \textbf{Avinash Sahu}$^{2}$,
    \textbf{Abdullah Mueen}$^{1}$,
    \textbf{Afsah Anwar}$^{1}$ \\ 
    $^{1}$Department of Computer Science, The University of New Mexico \\
    $^{2}$Comprehensive Cancer Center, The University of New Mexico \\
    \texttt{\{ajararweh,mikethebos,afsah\}@unm.edu}
}


\maketitle

\begin{abstract}
Increasing complexity in software systems places a growing demand on reasoning tools that unlock vulnerabilities manifest in source code. Many current approaches focus on vulnerability analysis as a classifying task, oversimplifying the nuanced and context-dependent real-world scenarios. Even though current code large language models (LLMs) excel in code understanding, they often pay little attention to security-specific reasoning. We propose LLaVul, a multimodal LLM tailored to provide fine-grained reasoning about code through question-answering (QA). Our model is trained to integrate paired code and natural queries into a unified space, enhancing reasoning and context-dependent insights about code vulnerability. To evaluate our model performance, we construct a curated dataset of real-world vulnerabilities paired with security-focused questions and answers. Our model outperforms state-of-the-art general-purpose and code LLMs in the QA and detection tasks. We further explain decision-making by conducting qualitative analysis to highlight capabilities and limitations. By integrating code and QA, LLaVul enables more interpretable and security-focused code understanding.\\
\end{abstract}


\section{Introduction}


As systems grow in size and complexity, context-dependent flaws emerge and enable opportunities for critical security vulnerabilities. These vulnerabilities can have led to malicious attacks, impacting the Confidentiality, Integrity, and Availability (CIA) properties of their systems. Additionally, the vulnerabilities can have varying effects on the systems, including arbitrary code execution on systems \cite{gonçalves2025evaluatingllama32software}. Understanding and identifying security vulnerabilities is deemed to be paramount to reducing potential exploitation. In practice, this process requires a deep understanding of the development environment, programming language, availability of proper documentation, and significant manual inspection and testing of different use cases \cite{zhang2024}. In time-sensitive environments, this process could be time-consuming and resource-intensive. Further, the number of publicly disclosed vulnerabilities every year is continuously piling on \cite{anwar2021cleaning,anwar2022recent}.
Therefore, automated approaches have been proposed to mitigate these challenges by scanning the code to detect security vulnerabilities or weaknesses \cite{inproceedings,staticanalysistools}. However, these tools often depend on heuristics, predefined rules, and lack the understanding of code semantics, limiting their generalizability to emerging vulnerabilities \cite{li2024empirical,nguyen2021ranking}. \\

To enable deeper understanding of code semantics and better generalizability, machine-learning approaches have been introduced to learn from large-scale corpora of code and their known vulnerabilities. 
These efforts focused mainly on classification tasks in a given code and spanned two classification settings, such as vulnerable or non-vulnerable \cite{VulDeePecker, Devign, Vul-LMGNN}. Prior works have also explored multi-class classification tasks to detect vulnerability types (e.g., overflows) in vulnerable sources \cite{muVulDeePecker, REVEAL, SySeVR, VulCNN}. To do so, the models depended on features derived from syntax trees \cite{VulDeePecker, Devign}, Control Flow Graphs (CFGs) \cite{SySeVR, REVEAL, Vul-LMGNN}, or learned code embeddings from foundational models \cite{SecureBERT, VulBERTa}. While these approaches have advanced the field, they often oversimplify the problem space by treating it as a classification task. For example, code snippets that contain multiple security vulnerabilities may be underrepresented by predicting a single vulnerability, leaving the system prone to exploitation. Similarly, these models usually fail to provide insights into where these vulnerabilities are located or how they can be solved.

On the other hand, Large Language Models (LLMs) offer a new paradigm for reasoning in a variety of fields. Specifically, LLMs trained on massive corpora of code QA have been shown to possess a rich semantic understanding of code and support more interpretable output \cite{zhang2024}. Unlike classification models, LLMs operate as copilots and can engage in thorough conversations, providing deeper insights and fine-grained contextual explanations. Recent advances in training paradigms and benchmarks of open-source code LLMs \cite{OpenCoder, qwen2, CodeT5, CodeLlama, fried2022incoder} have focused their attention on general code understanding tasks with little awareness of security context \cite{basic2025largelanguagemodelscode, jiang2025investigatinglargelanguagemodels}. On the other hand, current code vulnerability datasets/benchmarks are usually limited to binary and multi-class classification problems \cite{Devign, DiverseVul, CodeXGLUE, SySeVR, Vul-LMGNN, VulDeePecker}, and lack detailed explanations or fine-grained reasoning of vulnerabilities. Thus, it underscored the need for more comprehensive benchmarks or generation pipelines reflecting the nature of software development vulnerabilities.

Therefore, we propose LLaVul: \textbf{L}arge \textbf{L}anguage \textbf{a}nd \textbf{Vul}nerability assistant, a novel multimodal LLM inspired by LLaVA architecture that integrates source code with security-focused QA. A software engineer could use our model to gain insights into known vulnerabilities during development, potentially preventing end-users from being exposed to security flaws. Our model is designed to reason about code vulnerability in an interpretable manner by leveraging code semantics and natural language. More specifically, LLaVul is trained on pairs of QA and code segments to answer security-driven questions. Furthermore, we introduce LLaVul-QA, a new benchmark tailored for vulnerability QA, spanning real-world code programs paired with questions and answers generated from the CVEFixes and MSR dataset \cite{CVEfixes, MSR}. This benchmark is designed to advance currently proposed datasets for training security-focused LLMs beyond classification problems and general code understanding tasks. We comprehensively assess our model performance on our curated QA dataset and DiverseVul \cite{DiverseVul} classification benchmark. We also conduct ablation and qualitative analysis to highlight model capabilities and limitations.

The remainder of this paper is organized as follows: Section~\ref{sec:rw} explores related work in the literature. Section~\ref{sec:method} presents our proposed framework for vulnerability QA, and Section~\ref{sec:results} highlights our experimental setup and reports quantitative results. Section~\ref{sec:discussion} discusses practical implications and proposes future directions, and Section~\ref{sec:conclusion} concludes with a summary of our contributions.

\section{Related Work}
\label{sec:rw}

\subsection{General Large Language Models Pretraining} Following the introduction of the self-attention mechanism in Transformers, Large Language Models (LLMs) capabilities were significantly enhanced in understanding long-range dependencies \cite{attention}. LLMs utilizing Transformer architecture have been widely applied to perform various tasks. For example, BERT variants LLMs are encoder-based architectures trained to predict the masked token/sentence in a corpus of text \cite{bert, roberta}. These models are usually fine-tuned to perform downstream tasks such as classification. On the other hand, encoder-decoder and decoder-only LLMs mainly focus on autoregressive text generation tasks \cite{llama2, gpt3}. Further training paradigms were introduced, such as instruction tuning, where the models are fine-tuned to follow natural language instruction, enabling zero-shot capabilities on a wide range of tasks \cite{vicuna, gpt4}. Instruction tuning has also been applied in multimodal contexts where the model is instructed to perform cross-modal tasks such as multimodal QA \cite{llava}. 

\subsection{LLMs for Programming Languages} The evolution of code-based LLMs started with encoder-based LLMs, which attempt to find rich code representations  \cite{CodeBERT, GraphCodeBERT, CodeSage}. Once fine-tuned, these models demonstrate remarkable capabilities in code understanding tasks such as code search. Limited by their architectural design, these models inherently lack generative capabilities. Thus, encoder-decoder and decoder-only models were introduced to enable more code-related tasks such as code generation, code summarization, code translation, etc \cite{AlphaCode, CodeT5, PolyCoder}. Since then, generative models have been adapted to different software engineering tasks, enhancing programming capabilities \cite{zhang2024unifyingperspectivesnlpsoftware}.

\subsection{LLMs for Security and Vulnerability Analysis} 
Recent work has also explored the applications of LLMs in different security contexts. For example, Chen et al. investigated the ability of LLMs to address security and privacy misconceptions by prompting engineering, indicating LLMs' ability to serve as an automated advisor \cite{chen2023largelanguagemodelsprovide}. Most recently, Li et al. explored localizing the security vulnerability in code at the line level by tracking attention weights inherent in LLMs \cite{li2024attentionneedllmbasedcode}. Combined with traditional methods, such as search engines and heuristic-based text matches, LLMs have also been adopted to find vulnerabilities at the file level from natural language vulnerability reports \cite{Sun2024}. Other research efforts attempted to identify security vulnerabilities in source code. Different models utilized modern deep learning architectures to detect security vulnerabilities, such as multilayer perceptron \cite{aibughunter}, LSTM-based \cite{VulDeePecker,muVulDeePecker}, and graph-based \cite{Devign, Vul-LMGNN, REVEAL, SySeVR}. Utilizing recent advances in LLMs, downstream fine-tuning of BERT model variants has also been adopted to detect security vulnerabilities in code \cite{SecureBERT, VulBERTa, LProtector}. However, these attempts were mainly tailored to vulnerability classification tasks. Generative LLMs for code, such as CodeT5 \cite{CodeT5}, and AlphaCode \cite{AlphaCode}, could be utilized to produce generative output that describes security vulnerabilities \cite{LLM4Vuln}. However, these models are trained on general code and not tailored to security vulnerability tasks.



\begin{table}
\centering
\normalsize
\begin{tabular}{lll}
\toprule
Dataset & Pretraining & fine-tuning \\
\midrule
\# codes & 549708 & 104977 \\
\# CVEs & - & 6760 \\
\# QAs & 549708 & 406885 \\
max \# QAs per code & 2 & 11 \\
min \# QAs per code & 1 & 1 \\
mean \# QAs per code & 1 & 3.87 \\
\# tokens & 78117083 & 60292936 \\
max \# tokens per code & 13041 & 11053 \\
min \# tokens per code & 7 & 3 \\
mean \# tokens per code & 142.11 & 574.34 \\
\bottomrule
\end{tabular}
\caption{QA pair and token counts for the pretraining and QA fine-tuning datasets.}
\label{tab:stats}
\end{table}

\section{Methodology}
\label{sec:method}

In this section, we introduce a collection of pretraining code snippets and vulnerable code, with their detailed descriptions. Then, the LLaVul model is introduced with details on pretraining and fine-tuning.

\subsection{Datasets} 
We collect two datasets to train our model: pretraining and fine-tuning datasets.

\textbf{Pretraining dataset.} The pertaining dataset was collected from the CodeXGLUE benchmark \cite{CodeXGLUE} for code summarization. The dataset contains programming language functions and their descriptions. We convert this dataset to a QA dataset that fits the model input format. We use the following prompt as a question \textit{"Summarize the purpose of this script."}, the text summary as the response, and functions from the benchmark as input to the code modality. This dataset is mainly used to train the projector to map from code token embeddings to the text token space. Table~\ref{tab:stats} demonstrates a high-level overview of the generated dataset.\\

\textbf{Fine-tuning dataset: LLaVul-QA.} 
We first utilized the open-source vulnerable code snippets and detailed descriptions of the vulnerabilities in the code. We collected around 160,000 vulnerable functions from the CVEfixes \cite{CVEfixes} and MSR \cite{MSR} datasets. C, C++, and JavaScript were the most common languages of code containing vulnerabilities, primarily due to an abundance of memory management errors and the ubiquity of web applications, respectively. Next, we use the open-source `Llama-3.1-8B-Instruct' model to generate the QA data by supplying pairs of source codes and their code/vulnerability descriptions. The model was prompted to simulate a conversation between a human and an assistant about the provided code and summaries, to extract information related to program semantics and vulnerability location, fix, mechanism of action, severity, etc. Table~\ref{tab:stats} demonstrates high-level overview statistics of the QA dataset, and Table~\ref{app_tab:LLaVul-QA-sample} exhibits a sample conversation from our dataset. Moreover, the model was specifically prompted to generate questions and answers based on the descriptions and code without introducing extra information. In practice, this process is best performed by humans, but it is often labor-intensive and expensive. As a result, we follow recent studies that have focused on developing automated solutions for similar tasks \cite{llava,DeepSeek_R1}. \\


\begin{figure}
    \centering
    \includegraphics[width=0.9\linewidth]{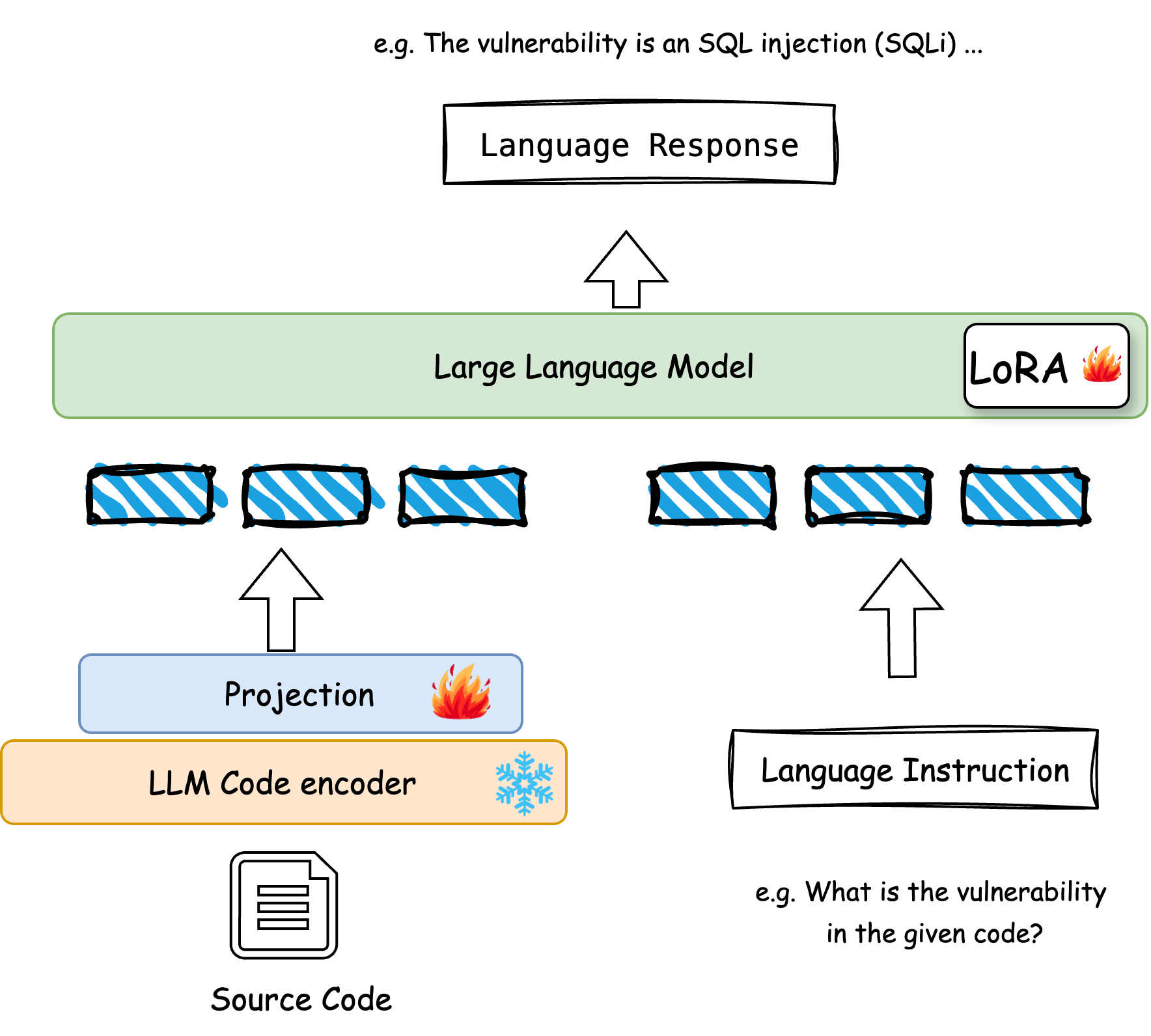}
    \caption{\textbf{High Overview of LLaVul}. LLaVul maps the source code tokens to the language space (denoted by blue rectangles) and combines them with the instruction tokens to generate security insights about the source code. During the fine-tuning stage, the trainable model parts are denoted by the fire symbol (\includegraphics[width=0.02\textwidth]{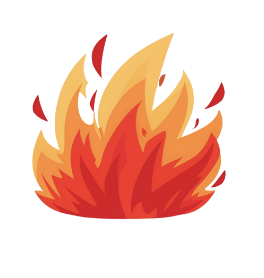}) (where the LLM is trained using low rank matrices via LoRA) while the frozen parts are denoted by the snow symbol (\includegraphics[width=0.02\textwidth]{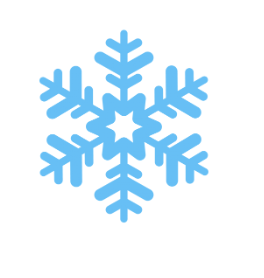}). }
    \label{fig:archi}
\end{figure}

\subsection{LLaVul Architecture}

\textbf{Code Encoder.}
Our approach is based on LLaVA \cite{llava}, which integrates images and text via instruction tuning. We adapt LLaVA to source code snippets by replacing the image encoder with a code encoder (Figure~\ref{fig:archi}). We use CodeSage-small \cite{CodeSage}, a transformer-based encoder with 6 layers and 130 million parameters. Every snippet $(\mathcal{S})$ is encoded to a multidimensional token embedding using CodeSage-small, $\phi_{cs}$. Formally,
\begin{equation*}
    \mathbf{Z}_v=\phi_{cs}(\mathcal{S})
\end{equation*}
where $\mathbf{Z}_v \in \mathbb{R}^{d \times T_1}$ represents the embedding of the code tokens where $d$ is the dimension size and $T_1$ is the number of tokens.\\

\textbf{LLM Encoder.}
Simultaneously, the instruction/question $\mathbf{X}_q$, given as natural language input, is tokenized and encoded using a 7 billion parameter Vicuna 1.5 model \cite{vicuna}, $\phi_{LLM}$:
\begin{equation*}
    \mathbf{H}_q = \phi_{LLM} (\mathbf{X}_q)
\end{equation*}
where $\mathbf{H}_q\in \mathbb{R}^{k \times T_2}$ represents the token embeddings of the instruction where $k$ is the embedding dimension and $T_2$ is the number of tokens. \\

\textbf{Code Projector.}
To align the source code and text modalities (Figure~\ref{fig:archi}), we project the code latent tokens (each of dimension $d$) into the language embedding space (dimension $k$) via the projector:
\begin{equation*}
    \mathbf{H}'_v = W \cdot \mathbf{Z}_v
\end{equation*}
where $W$ is the set of trainable parameters and $\mathbf{H}'_v \in \mathbb{R}^{k \times T_1}$. The projected tokens are then concatenated to the text tokens, producing $\mathbf{H}\in \mathbb{R}^{k \times \left(T_1+T_2\right)}$. $\mathbf{H}$ is then fed into the LLM decoder, $f_\phi$, to generate the response.\\

\textbf{Training.}
The model training process consists of two stages: pretraining and fine-tuning. During pretraining, we freeze the code encoder and the LLM while the parameters for the projector are trained. The main objective of this stage is to enable the model to learn the projection from the code token space to the text space. Then, we perform fine-tuning, where we train the entire model except the code encoder parameters on a downstream task such as question-answering and vulnerability classification. In this stage, the LLM is trained using Low-Rank Adaptation (LoRA) \cite{lora}. Table~\ref{app_tab:hyperparameters} illustrates the hyperparameters used during the two training stages. 

Model training and inference were mainly performed on 2 to 4 NVIDIA A100 PCIe GPUs of 40GB VRAM, respectively. The estimated training time is roughly dependent on the number of parameters, the batch sizes, and other configurations such as gradient checkpointing and LoRA parameters. However, the estimated training time for the pretraining stage is around 5 days, while the fine-tuning stage varies from 12 to 72 hours. The generation of QA pairs to form the fine-tuning dataset took multiple weeks on a variety of resources, including 4 A100s, V100s, and an A6000. The time estimations are based on the parameters found in Table~\ref{app_tab:hyperparameters}. The table also highlights the best-performing model parameters of the experiments in this manuscript.\\

\begin{table*}
\centering
\small
\begin{tabular}{lccccccc}
\hline
\textbf{Model} &
\makecell{\textbf{BLEU}\\\textbf{2}} &
\makecell{\textbf{BLEU}\\\textbf{4}} &
\makecell{\textbf{ROUGE}\\\textbf{1}} & \makecell{\textbf{ROUGE}\\\textbf{2}} & \makecell{\textbf{ROUGE}\\\textbf{L}} & \makecell{\textbf{METEOR}\\\textbf{}}&
\makecell{\textbf{BERT}\\\textbf{Score}}
\\ \midrule

\rowcolor[HTML]{bdcbff} 
\multicolumn{8}{l}{\textbf{\textit{General-purpose LLMs}}} \\

DeepSeek-R1-Distill-Llama-8B & 0.011 & 0.004 & 0.041 & 0.0169 & 0.034 & 0.092 & 0.689  \\

DeepSeek w/o thinking & 0.021 & 0.007 & 0.127 & 0.037 & 0.097 & 0.190 & 0.709 \\

Qwen2-7B-Instruct & 0.016 & 0.005 & 0.076 & 0.022& 0.057 & 0.140 & 0.675  \\
\rowcolor[HTML]{bdcbff} 
\multicolumn{8}{l}{\textbf{\textit{Code-specific LLMs}}} \\

OpenCoder-8B-Instruct & 0.013 & 0.005 & 0.121 & 0.048 & 0.098 & 0.193 & 0.734  \\

Qwen2.5-Coder-7B-Instruct & 0.018 & 0.006 & 0.094 & 0.027 & 0.070 & 0.163 & 0.697  \\

CodeLlama-7b-Instruct & 0.032 & 0.015 & 0.141 & 0.066 & 0.120 & 0.211 & 0.705  \\

CodeT5-base-multi-sum & 0.043 & 0.017 & 0.199 & 0.048 & 0.161 & 0.129 & 0.821 \\

LLaVul (Ours) & \textbf{0.267} & \textbf{0.168} & \textbf{0.390} & \textbf{0.234} & \textbf{0.350} & \textbf{0.353} & \textbf{0.910}  \\

\hline
\end{tabular}
\caption{\textbf{Baseline comparison on our LLaVul test set}. \textbf{Bold} and \underline{underline} denote best and second best performing models respectively. For all metrics, higher values indicate better performance.}
\label{tab:baseline_results}
\end{table*}

\section{Results}
\label{sec:results}


\subsection{LLaVul QA}
\label{sec:results_LLaVul_QA}

 \textbf{Experiment.} 
We assessed LLaVul's performance by comparing it to two classes of LLMs: general-purpose LLMs and code-specific LLMs. For the general-purpose category, we choose the Llama distilled DeepSeek model \cite{DeepSeek_R1}. We provide two predictions with and without adding the thinking-generated text. We also utilize the Qwen2 model \cite{qwen2}. Next, we compare to LLMs that are specifically tailored for code QA tasks: OpenCoder \cite{OpenCoder}, QwenCoder \cite{qwen2}, CodeLLaMA \cite{CodeLlama}, and CodeT5 \cite{CodeT5}. For all LLMs in this experiment, we utilized open-source weights and inference of the models by passing both the question and the code script. We evaluated the performance using lexical metrics such as BLEU \cite{bleu}, ROUGE \cite{ROUGE}, and METEOR \cite{METEOR}, and semantic similarity metrics such as BERT similarity \cite{bert}. BLEU and ROUGE measure F1 scores using the longest common subsequence of tokens, while METEOR is more flexible, comparing multiple chunks of common subsequences. \\ 


\textbf{Findings.}  Table~\ref{tab:baseline_results} demonstrates the performance of the baselines on LLaVul QA.  First, general-purpose LLMs showed lower performance since they are tailored to more general domains and not trained on code QA. For example, Qwen2 and QwenCoder models stem from the same LLM architecture, with QwenCoder being fine-tuned on code QA using instruction tuning. The results showed that QwenCoder outperformed the general Qwen2 model across all metrics, showcasing the value of code instruction-tuning.  Similarly, the results also showed that all other code-specific LLMs demonstrated relatively better performance than general-purpose LLMs in answering code vulnerability questions. This is expected since these models are trained on large-scale datasets, allowing them to understand code structure more effectively. \\

Finally, LLaVul outperformed the proposed baseline LLMs from both categories across the semantic and lexical metrics. However, models like CodeT5 showed a competitive BERT score (semantic similarity to the ground truth), which might suggest that the answers are relevant but not necessarily accurate (see Table~\ref{tab:QA_qualitative}). Consistent with the previous argument, we hypothesized that the baseline LLMs may show lower performance due to being evaluated on a newly curated benchmark with cross-domain data or due to their lack of focus on code vulnerability QA, suggesting potential challenges in generalizability.

\begin{table}[ht]
    \small
    \centering
\begin{tabular}{l l c c c}
    \toprule
    \textbf{Model} & \textbf{Accuracy} & \textbf{F1 Score} & \textbf{Precision} & \textbf{Recall} \\
    \midrule

    \rowcolor[HTML]{DFFFD6} 
    \multicolumn{5}{l}{\textbf{\textit{Encoder-decoder LLMs}}} \\
    CodeT5       & 0.47 & 0.36 & 0.48 & 0.47 \\
    LLaVul (Ours)& 0.52 & 0.48 & 0.55 & 0.52 \\

    \rowcolor[HTML]{DFFFD6} 
    \multicolumn{5}{l}{\textbf{\textit{Encoder-only Foundation Models}}} \\   
    CodeBERT      & 0.72 & 0.69 & \textbf{0.77} & 0.62 \\
    GraphCodeBERT & 0.74 & 0.74 & 0.74 & 0.74 \\
    CodeBERTa     & 0.74 & 0.76 & 0.71 & 0.81 \\
    LLaVul (Ours) & \textbf{0.75} & \textbf{0.77} & 0.71 & \textbf{0.84} \\
    \bottomrule
\end{tabular}
\caption{\textbf{Vulnerability classification on the DiverseVul benchmark}. Performance comparison of generative (encoder-decoder) LLMs and foundation models (i.e., encoder-only) on the vulnerability classification task. \textbf{Bold} denotes the best-performing model. For all metrics, higher values indicate better performance.}
\label{tab:vul_classification}
\end{table}

\begin{table*}
\centering
\resizebox{\textwidth}{!}{
\begin{tabular}{p{2.5cm}|p{3cm}|p{16cm}}
\hline
\textbf{ID/Code/Question} & \textbf{Ground Truth} & \textbf{Model Outputs} \\ \hline

\textbf{ID}: CVE-2017-12595 \newline
\textbf{Code}: Table~\ref{app_tab:cvefixes18844} \newline
\textbf{Q}: What is the impact of the vulnerability in the given code?&
The impact of this vulnerability is a denial of service (stack consumption and segmentation fault) or possibly unspecified other impact. & 
\textbf{LLaVul:} This vulnerability allows remote attackers to cause a \textcolor{blue}{denial of service (buffer under-read and application crash)} or possibly have unspecified other impact \textcolor{teal}{via a crafted document.} \newline
\textbf{Deepseek:} I identified vulnerabilities in the QPDF code primarily revolve around potential issues with \textcolor{blue}{buffer management}, stream handling, and loop termination. \newline
\textbf{CodeT5:} Parse a content stream of a single object. \newline
\textbf{CodeLLaMA:} The impact of this vulnerability is that it allows an attacker to inject arbitrary code into a PDF file, \textcolor{teal}{potentially leading to arbitrary code execution on the system where the PDF is opened.} The vulnerability is caused by the `parseInternal` function, which takes a `Buffer` object as input and parses it as a PDF file. \newline
\textbf{OpenCoder:} The impact of this vulnerability is that it can lead to a \textcolor{blue}{Denial of Service (DoS) attack}. This is because the vulnerability \textcolor{teal}{allows an attacker to input a specially crafted PDF file that causes the system to crash} or become unresponsive when trying to process the file. \\  \hline

\textbf{ID}: CVE-2022-3138 \newline
\textbf{Code}: Table~\ref{app_tab:cvefixes83065} \newline
\textbf{Q}: What is the type of vulnerability identified in the given code?&
The vulnerability identified in the given code is a Cross-site Scripting (XSS) vulnerability.& 
\textbf{LLaVul:} \textcolor{blue}{The vulnerability is a Cross-site Scripting (XSS) }- Generic. XSS vulnerabilities can be exploited to inject malicious scripts into a web page, potentially allowing an attacker to steal user data...\newline
\textbf{Deepseek:} The code exhibits a possible information leakage or configuration vulnerability in the `isPagesEnabled` method, where pages may be enabled unintentionally if the `hide-pages` URL parameter isn't set, potentially exposing sensitive data. \newline
\textbf{CodeT5:} This is the type of vulnerability that should be displayed in the editor. \newline
\textbf{CodeLLaMA:} vulnerability identified in this code is a type of injection vulnerability, specifically a \textcolor{blue}{reflected cross-site scripting (XSS) vulnerability}. The vulnerability arises from the use of user-supplied data in the `createTemporaryGraph` method, which is then used to create a new graph object... \newline
\textbf{OpenCoder:} The type of vulnerability identified in the given code is not specified. However, it seems to be a logical error or flaw ... \\  \hline

\textbf{ID}: CVE-2021-27341 \newline
\textbf{Code}: Table~\ref{app_tab:cvefixes35496} \newline
\textbf{Q}: Is there any specific code snippet related to a vulnerability?&
No, there is no specific code snippet related to a vulnerability in the provided source code.& 
\textbf{LLaVul:} The provided code snippet \textcolor{blue}{is not related to a vulnerability}. \underline{The vulnerability is in the framework} \underline{ helpController.php file, which is not provided}. \newline
\textbf{Deepseek:} The potential vulnerability lies in how keys are being managed and processed... \newline
\textbf{CodeT5:} This method checks if there any code snippet related to this vulnerability. \newline
\textbf{CodeLLaMA:} The code snippet you provided is a function that takes an array as input and returns an array with the same values, but with all keys converted to uppercase ... The vulnerability you mentioned is a type of injection vulnerability, specifically a reflected XSS (cross-site scripting) vulnerability. \newline
\textbf{OpenCoder:} Your code snippet seems to be a function that is intended to clean up and reformat an array. \textcolor{blue}{However, it does not seem to be directly related to a specific vulnerability.} \\ 
\bottomrule
\end{tabular}
}
\caption{\textbf{Qualitative analysis: Vulnerability QA.} Analyzing baselines performance on vulnerability-related questions. \textcolor{blue}{Blue} color indicates high relevance to the ground truth. \textcolor{teal}{Teal} color indicates pieces of text relevant to the code.  \underline{Underline} showcases an example hallucination of our model. 
}
\label{tab:QA_qualitative}
\end{table*}

\subsection{Vulnerability Classification: DiverseVul}

 \textbf{Expirement.} We evaluate our model performance on the DiverseVul benchmark \cite{DiverseVul}. This benchmark is dedicated to vulnerability classification and spans 18,945 vulnerable and 330,492 non-vulnerable functions of 150 different CWEs extracted from GitHub commits.  We balance the dataset by randomly sampling 18,945 non-vulnerable functions. We compare the performance of our model across generative LLMs: CodeT5 \cite{CodeT5}, and foundation models designed for code understanding tasks: CodeBERT \cite{CodeBERT}, GraphCodeBERT \cite{GraphCodeBERT}, and CodeBERTa \cite{CodeBERTa_not}. The foundation models in this experiment were fine-tuned specifically to predict the potential vulnerability of given source codes. \\

 \textbf{Findings.} Table~\ref{tab:vul_classification} demonstrate the results on the DiverseVul benchmark. We first compared the performance of our model to CodeT5 in predicting vulnerable functions. Both models exhibited a lower performance, with LLaVul being the best. Due to the nature of their training and architectural design, even though generative LLMs show advanced capabilities on a variety of tasks, they may not always outperform specialized models like BERT in classification tasks \cite{LLM_classification1, LLM_classification2, LLM_classification3}. To this end, we fine-tuned the encoder part of our model and compared it to fine-tuned BERT-based models in this task. Our encoder-based model achieved the performance of other encoder-based models, suggesting that generative LLMs might not always be suitable for classifying vulnerable functions without prior model engineering.

\subsection{Qualitative Analysis}

Table~\ref{tab:QA_qualitative} demonstrates a few sample predictions of the baselines in Section~\ref{sec:results_LLaVul_QA}. LLaVul shows relevant responses that often align with the ground truth. The inference ability of our model not only covers the text modality (i.e., the prompt, highlighted in \textcolor{blue}{Blue}) but also expands to the code modality (highlighted in \textcolor{teal}{Teal}), indicating successful projection from code to text modality. We also noticed that general-purpose LLMs such as DeepSeek usually tend to find the root cause of the problem by analyzing the code piece by piece, and then draw conclusions based on this analysis, which might need to be verified manually to see if they align with the CVEFixes descriptions. On the other hand, Code LLMs such as CodeT5 and OpenCoder showed lower generalizability to new prompts, and their answers were related to the code alone, ignoring the paired question. Finally, our model exhibits some hallucinations as indicated by the underlined text in example id `CVE-2021-27341'. More specifically, the model admits that the file 'helpController.php' is not provided; however, it still predicts that the vulnerability is coming from that file, suggesting more room for improvement.  During hallucinations, many of the specific filenames and functions mentioned by the model can be found specifically in our fine-tuning datasets, suggesting that the model could feature some overfitting to the data. \\


\begin{table}[H]
\centering
\small
\begin{tabular}{lcccc}
\hline
\textbf{LLaVul} &
\makecell{\textbf{BLEU}\\\textbf{4}} &
\makecell{\textbf{ROUGE}\\\textbf{L}} &
\makecell{\textbf{METEOR}\\\textbf{}} &
\makecell{\textbf{BERT}\\\textbf{Score}} \\
\hline

\textbf{Only LLM} & 0.082 & 0.287 & 0.313 & 0.886 \\
~\textit{w/ LLaMA3.1} &&&& \\

\textbf{Only Pretraining} & 0.004 & 0.045 & 0.080 & 0.627 \\

\textbf{Finetuned} & 0.031 & 0.109 & 0.249 & 0.729 \\
~\textit{100 Code Tokens} &&&& \\

\textbf{Finetuned} & \textbf{0.168} & \textbf{0.350} & \textbf{0.353} & \textbf{0.910} \\
~\textit{Full Tokens} &&&& \\

\hline
\end{tabular}
\caption{\textbf{Ablation study}. Denoted in \textbf{Bold} is the best-performing model. Code length refers to the number of code tokens passed along the prompt. We show fewer metrics for compactness. Refer to Table~\ref{app_tab:ablation_rest} for full metrics.}
\label{tab:ablation}
\end{table}

\subsection{Ablation Study}
Since large-scale ablation studies are prohibitive in LLMs due to their resource need and time-consuming nature, we focus on targeted ablation studies to evaluate major parts of the multimodal architecture. The results reported in this experiment are based on the LLaVul QA. We first assess the importance of the code modalities by reducing the number of code tokens passed to LLaVul over time. The results show that increasing the number of tokens beyond 100 code tokens demonstrates better performance (Table~\ref{tab:ablation}). Note that the average code token length in the QA data is approximately 574 code tokens (Table~\ref{tab:stats}). Next, we evaluate the importance of the fine-tuning stage beyond pertaining. Table~\ref{tab:ablation} demonstrates superior performance of fine-tuned LLaVul, showcasing the essentiality of fine-tuning. Lastly, we evaluate code instruction tuning as a multimodal setting by comparing it to LLaMA3.1 LLM (single-modality LLM). We pass the code and the text prompt in the same query to LLaMA and evaluate the response. Yet competitive, LLaVul with code instruction tuning (i.e., LLaVul - Finetuned w/ Full code length) showed enhanced performance on the vulnerability QA task.\\

\section{Discussion}
\label{sec:discussion}
Our experiments demonstrate that the multimodal design of LLaVul, which integrates code and natural language, yields a consistent gain over the proposed baselines in open-ended vulnerability QA. On classification tasks such as DiverseVul, we demonstrate that our encoder-based model achieves equivalent performance or often outperforms discriminative transformers like CodeBERT and GraphCodeBERT, showing the ability to provide both binary vulnerability signals and interpretable text. The qualitative analysis and ablation studies focus on exploring the model strengths and weaknesses, as well as providing insight into the model design choices.

While LLaVul represents a step toward explainable, context-aware vulnerability analysis, several avenues remain open. One avenue is to mitigate model hallucinations by integrating human-in-the-loop validation or a second LLM agent's verification of the generated QA. This would ensure that the learned behaviour during the fine-tuning is reliable for safety-critical deployments and applicable to a wide range of scenarios. This work presents a step forward for converting raw data into actionable human-like conversations. However, future work should explore better utilization and refinement of the abundantly available code vulnerability information. Finally, current approaches are still lacking due to their focus on single-file vulnerability identification; thus, extending this problem to multi-file schemes could further boost precision and recall, especially for rare CWEs.

\section{Conclusion}
\label{sec:conclusion}

We proposed LLaVul, a multimodal LLM tailored for open-ended security-focused QA in code. By integrating code and QA into a single language space, our model goes beyond standard approaches that tackle the problem as a classification task by enabling context-dependent reasoning and interpretable vulnerability analysis. We also introduced a new security-focused QA paired with source code from real-world software vulnerabilities. Our analysis showed that our model outperforms general-purpose and code LLMs in the constructed security-based QA, indicating the need for more awareness of security understanding tasks. Our model also showed enhanced performance on the DiverseVul task compared to fine-tuned BERT-based foundation models. Furthermore, we conducted qualitative analyses on vulnerability QA and code summarization to demonstrate its ability to explain and reason about vulnerability by utilizing information from text and code modalities. By bridging the gap between code and security QA, LLaVul goes beyond traditional code LLMs and paves the way for more explainable, context-aware, and interactive AI tools in cybersecurity, such as incorporation into Integrated Development Environments (IDEs) for assisting programmers with maintaining secure code or automating useful alerts for upkeep of legacy programs with security fixes that require knowledge of the code's semantics.\\


\section{Acknowledgments}

We would like to thank the UNM Center for Advanced Research Computing, supported in part by the National Science Foundation, for providing the research computing resources used in this work.

\bibliographystyle{IEEEtran}
\bibliography{ref}

\appendices
\section{Appendix Tables}

\begin{table}[!t]
\centering
\large
\begin{tabular}{lcc}
\toprule
\textbf{Hyperparameter} & \textbf{Pre-training} & \textbf{Fine-tuning} \\
\midrule
\midrule
\rowcolor[HTML]{D6EAF8} 
\multicolumn{3}{l}{\textbf{\textit{Training}}} \\ 
Number of Epochs & 1 & 3 \\ 
Per-device Batch Size & 10 & 5 \\ 
Trinable parameters & 21M & 181M \\
Learning Rate & \multicolumn{2}{c}{$2 \times 10^{-5}$} \\ 
Precision & \multicolumn{2}{c}{bf16 (Mixed Precision)} \\
Optimizer & \multicolumn{2}{c}{AdamW}\\
\midrule
\rowcolor[HTML]{D6EAF8} 
\multicolumn{3}{l}{\textbf{\textit{Code Encoder}}} \\ 
Model & \multicolumn{2}{c}{codesage/codesage-small} \\ 
Output Tokens & \multicolumn{2}{c}{All (i.e. no truncation)} \\ 
Feature Layer & \multicolumn{2}{c}{-2 (i.e. second to last)} \\ 
Max Code Length & \multicolumn{2}{c}{1000} \\ 
\midrule
\rowcolor[HTML]{D6EAF8} 
\multicolumn{3}{l}{\textbf{\textit{Language Model}}} \\
Model & \multicolumn{2}{c}{lmsys/vicuna-7b-v1.5} \\ 
LoRA Rank & \multicolumn{2}{c}{64} \\ 
LoRA Alpha & \multicolumn{2}{c}{64} \\ 
Context Length & \multicolumn{2}{c}{2048} \\  
\midrule
\rowcolor[HTML]{D6EAF8} 
\multicolumn{3}{l}{\textbf{\textit{Projector}}} \\
Number of Layers & \multicolumn{2}{c}{2 layers} \\ 
Activation & \multicolumn{2}{c}{GELU} \\ 
Hidden Dimensions & \multicolumn{2}{c}{4096}\\
\bottomrule
\end{tabular}
\caption{An overview of the hyperparameters used to train the two stages of LLaVul. If one parameter is mentioned across the two columns, the same value is used in the two training stages. }
\label{app_tab:hyperparameters}
\end{table}

\begin{table*}[ht]
\centering
\small
\begin{tabular}{p{13cm}}
\hline
\multicolumn{1}{c}{\textbf{LLaVul-QA}} \\ \hline
\textcolor{blue!80}{Q1:  What is the vulnerability in this code?}  \\
\textcolor{red!50}{A1:  The vulnerability in this code is a heap use after a free bug in the PDF parser. It could allow an attacker to execute arbitrary code, read memory, or cause a denial of service via a crafted file.}  \\ 
\midrule
\textcolor{blue!80}{Q2:  Where in the code is the vulnerability located?}  \\
\textcolor{red!50}{A2:  The vulnerability is located in the add\_range function, specifically in the code that handles the insertion of new nodes into the splay tree.} \\ 
\midrule
\textcolor{blue!80}{Q3:  What is the mechanism of action for this vulnerability?} \\
\textcolor{red!50}{A3:  The mechanism of action is that an attacker can create a crafted PDF file that triggers the use after a free bug, allowing them to execute arbitrary code, read memory, or cause a denial of service.} \\ 
\midrule
\textcolor{blue!80}{Q4:  How can this vulnerability be fixed?} \\
\textcolor{red!50}{A4:  The vulnerability can be fixed by properly checking and handling the free use of memory in the add\_range function, and by ensuring that the splay tree is properly updated and maintained.}  \\ 
\midrule
\textcolor{blue!80}{Q5:  What is the severity of this vulnerability?}  \\
\textcolor{red!50}{A5:  The severity of this vulnerability is high, as it could allow an attacker to execute arbitrary code, read memory, or cause a denial of service.}  \\ 
\midrule
\end{tabular}
\caption{\textbf{Sample of our LLaVul-QA dataset.} The data is extracted for the CVE-ID CVE-2018-1000039 and C programming language.}
\label{app_tab:LLaVul-QA-sample}
\end{table*}

\begin{table*}
\centering
\small
\begin{tabular}{lccccccc}
\hline

\textbf{LLaVul} &
\makecell{\textbf{BLEU}\\\textbf{2}} &
\makecell{\textbf{BLEU}\\\textbf{4}} &
\makecell{\textbf{ROUGE}\\\textbf{1}} & \makecell{\textbf{ROUGE}\\\textbf{2}} & \makecell{\textbf{ROUGE}\\\textbf{L}} & \makecell{\textbf{METEOR}\\\textbf{}}&
\makecell{\textbf{BERT}\\\textbf{Score}}
\\ \midrule

\textbf{Only LLM} & 0.160 & 0.082 & 0.335 & 0.165 & 0.287 & 0.313 & 0.886  \\
~~ \textit{w/ LLaMA3.1}  &&&&&&&  \\

\textbf{Only Pretraining} & 0.014 & 0.004 & 0.053 & 0.013 & 0.045 & 0.080 & 0.627  \\

\textbf{Finetuned} & 0.051 & 0.031 & 0.125 & 0.070 & 0.109 & 0.249 & 0.729 \\
~\textit{100 Code Tokens}  &&&&&&&  \\

\textbf{Finetuned} & \textbf{0.267} & \textbf{0.168} & \textbf{0.390} & \textbf{0.234} & \textbf{0.350} & \textbf{0.353} & \textbf{0.910}  \\
~~ \textit{Full Tokens} &&&&&&&  \\

\hline
\end{tabular}

\caption{\textbf{Ablation study}. Denoted in \textbf{Bold} is the best-performing model. Code length refers to the number of code tokens passed along the prompt. }
\label{app_tab:ablation_rest}
\end{table*}

\begin{table*}[h!]
\centering
\begin{lstlisting}[language=C, basicstyle=\ttfamily\small, frame=single]
unset($return[$key]);
    }
}
break;
}
return @array_change_key_case($return, CASE_UPPER);
\end{lstlisting}
\caption{Source code for example id `CVE-2021-27341' in Table~\ref{tab:QA_qualitative}.
\label{app_tab:cvefixes35496}
}
\end{table*}

\begin{table*}[h!]
\small
\centering
\begin{lstlisting}[language=C, basicstyle=\ttfamily\small, frame=single]
QPDFObjectHandle::parseContentStream_internal(
    PointerHolder<Buffer> stream_data, std::string const& description,
    ParserCallbacks* callbacks)
{
    size_t length = stream_data->getSize();
    auto input = new BufferInputSource(description, stream_data.getPointer());
    QPDFTokenizer tokenizer; tokenizer.allowEOF();
    bool empty = false;

    while (static_cast<size_t>(input->tell()) < length)
    {
        auto obj = parseInternal(input, "content", tokenizer, empty, 0, 0, false, false, true);
        if (!obj.isInitialized()) break;
        callbacks->handleObject(obj);

        if (obj.isOperator() && obj.getOperatorValue() == "ID")
        {
            char ch, buf[4] = {0}; bool done = false; std::string inline_image;
            input->read(&ch, 1);
            while (!done)
            {
                if (input->read(&ch, 1) == 0)
                    throw QPDFExc(qpdf_e_damaged_pdf, input->getName(), "stream data",
                    input->tell(), "EOF found while reading inline image");

                inline_image += ch;
                memmove(buf, buf + 1, 3); buf[3] = ch;

                if (strchr(" \t\n\v\f\r", buf[0]) && buf[1] == 'E' &&
                buf[2] == 'I' && strchr(" \t\n\v\f\r", buf[3]))
                {
                    done = true;
                    input->seek(-3, SEEK_CUR);
                    for (int i = 0; i < 4 && !inline_image.empty(); ++i) 
                    inline_image.pop_back();
                }
            }
            callbacks->handleObject(QPDFObjectHandle::newInlineImage(inline_image));
        }
    }
}

\end{lstlisting}
\caption{Source code for example id `CVE-2017-12595' in Table~\ref{tab:QA_qualitative}.
\label{app_tab:cvefixes18844}
}
\end{table*}

\begin{table*}[h!]
\centering
\small
\begin{lstlisting}[language=C, basicstyle=\ttfamily\small, frame=single]
EditorUi.prototype.isPagesEnabled = function() {
    return this.editor.editable || urlParams["hide-pages"] !== "1"; };

EditorUi.prototype.createTemporaryGraph = function(a) {
    return Graph.createOffscreenGraph(a); };

EditorUi.prototype.addChromelessClickHandler = function() {
    var a = urlParams.highlight;
    if (a && a.length > 0) a = "#" + a;
    this.editor.graph.addClickHandler(a); };

\end{lstlisting}
\caption{Source code for example id `CVE-2022-3138' in Table~\ref{tab:QA_qualitative}.
\label{app_tab:cvefixes83065}
}
\end{table*}

\end{document}